\documentclass[10pt,twocolumn,letterpaper]{article}

\usepackage{iccv}              

\definecolor{iccvblue}{rgb}{0.21,0.49,0.74}
\usepackage[pagebackref,breaklinks,colorlinks,allcolors=iccvblue]{hyperref}
\usepackage{csquotes}
\usepackage{graphicx}
\usepackage{amssymb,enumitem}
\usepackage{amsmath}
\usepackage{commath}
\usepackage{multirow,array, bigdelim, makecell, booktabs}
\usepackage{array, caption, floatrow, tabularx}
\usepackage{float}
\usepackage{hyperref}
\usepackage{mathrsfs}
\usepackage{dsfont}
\usepackage{bbm}

\def\paperID{1386} 
\def\confName{ICCV}
\def\confYear{2025}

\DeclareRobustCommand{\eg}{\textit{e.g.}}

\DeclareRobustCommand{\ie}{\textit{i.e.}}

\DeclareRobustCommand{\method}{RefDense}
\DeclareRobustCommand{\char}{Charades}
\DeclareRobustCommand{\thum}{MultiTHUMOS}
\DeclareRobustCommand{\ent}{Action-Entity}
\DeclareRobustCommand{\mot}{Action-Motion}

\newcommand{\linkc}{\textcolor[rgb]{1.0,0.1,0.57}}
\DeclareRobustCommand{\blue}{\textcolor[rgb]{0.0,0.1,0.95}}
\newcommand{\cmark}{\checkmark}  
\newcommand{\xmark}{\text{\sffamily X}}  

\title{Reframing Dense Action Detection ({\method}): \\ A Paradigm Shift in Problem Solving \& a Novel Optimization Strategy}

\author{Faegheh Sardari$^{1}$
\and
Armin Mustafa$^{1}$
\and
Philip J. B. Jackson$^{1}$
\and
Adrian Hilton$^{1}$
\and
$^{1}$ Centre for Vision, Speech and Signal Processing (CVSSP)\\
University of Surrey, UK \\
{\tt\small \{f.sardari,armin.mustafa,p.jackson,a.hilton\}@surrey.ac.uk}
}

\begin{document}
\maketitle
\begin{abstract}
Dense action detection involves detecting multiple co-occurring actions while action classes are often ambiguous and represent overlapping concepts. We argue that handling the dual challenge of temporal and class overlaps is too complex to effectively be tackled by a single network. To address this, we propose to decompose the task of detecting dense ambiguous actions into detecting dense, unambiguous sub-concepts that form the action classes ({\ie}, action entities and action motions), and assigning these sub-tasks to distinct sub-networks. By isolating these unambiguous concepts, the sub-networks can focus exclusively on resolving a single challenge, dense temporal overlaps. Furthermore, simultaneous actions in a video often exhibit interrelationships, and exploiting these relationships can improve the method performance. However, current dense action detection networks fail to effectively learn these relationships due to their reliance on binary cross-entropy optimization, which treats each class independently. To address this limitation, we propose providing explicit supervision on co-occurring concepts during network optimization through a novel language-guided contrastive learning loss. Our extensive experiments demonstrate the superiority of our approach over state-of-the-art methods, achieving substantial improvements of {\bf 3.8\%} and {\bf 1.7\%} on average across all metrics on the challenging benchmark datasets, {\char} and {\thum}. \href{https://github.com/}{\underline{\linkc{Our code will be released upon paper publication}}}.
\end{abstract}    
\section{Introduction}
\label{sec:intro}

\begin{figure}[t]
  \includegraphics[width=1.0\linewidth]{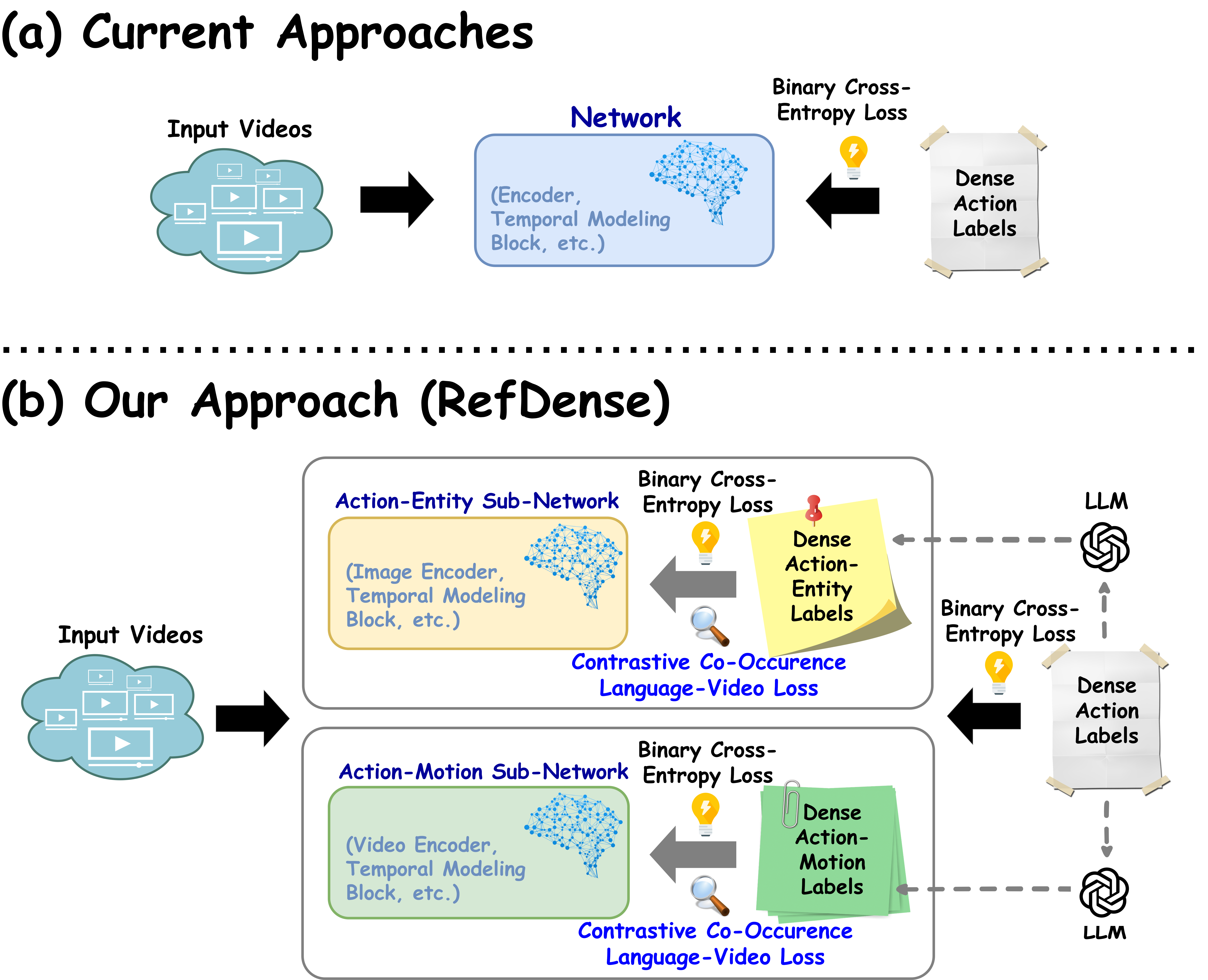}
  \caption{Comparison of current approaches and our proposed approach, {\method}, for tackling the dense action detection task. (a) Current approaches directly address the entire problem ({\ie}, detecting dense, ambiguous actions) using a single network, optimized solely with Binary Cross-Entropy (BCE) loss. In contrast, (b) {\method} decomposes the task into two sub-tasks ({\ie}, detecting dense, unambiguous entity and motion sub-concepts underlying the actions classes) and assigns them to distinct sub-networks. Furthermore, our approach is optimized using both BCE loss and our proposed contrastive co-occurrence language-video loss.}
  \label{fig: old vs ours}
\end{figure}

Dense action detection aims to recognize and temporally localize all actions within an untrimmed video, even when the actions occur concurrently. A deep understanding of these complex action semantics is crucial for many real-world applications, such as autonomous driving, sports analytics, and complex surveillance, where actions are rarely isolated. To tackle this task, current approaches \cite{dai2019tan, tirupattur2021modeling, dai2022ms, dai2023aan, sardari2023pat, zhu2024dual} typically follow a common pipeline. First, the video features are extracted using a pre-trained Encoder (e.g., I3D \cite{carreira2017quo}, CLIP \cite{radford2021learning}). Then, the features are fed into a Temporal Modeling Block (e.g., multi-scale transformer \cite{dai2022ms, sardari2023pat}, GNN \cite{dai2021ctrn, dai2023aan}) to capture temporal relationships among the segments, followed by a classification head that maps the learned representations to multi-action probabilities, enabling dense action detection. Finally, the entire network is optimized using binary cross-entropy (BCE) loss.

In dense action detection, beyond the challenge of temporal overlaps, action classes often exhibit semantic overlap ({\ie}, class ambiguity). This overlap can arise from shared entities or motion that define the action classes. For example, in the {\thum} dataset \cite{multithomus}, the action classes \enquote{Hammer Throw Wind Up} and \enquote{Hammer Throw Spin} share an identical entity, a hammer. Similarly, in the {\char} dataset \cite{charades}, the classes \enquote{Holding a Bag} and \enquote{Holding a Sandwich} overlap in motion, the act of holding. We argue that the dual challenge of handling temporal and action class overlaps is too complex to be effectively addressed by the traditional dense action detection pipeline. This motivated us to raise a novel question: {\bf Can we reduce the problem's complexity by eliminating the class overlaps, thereby enabling the network to focus solely on resolving temporal overlaps?} To achieve this, we introduce a paradigm shift in solving this task. Instead of directly detecting dense, ambiguous actions using a single network, we propose decomposing the task into detecting dense, unambiguous sub-concepts underlying the action classes ({\ie}, entity and motion sub-concepts), and assigning these sub-tasks to distinct sub-networks. By isolating the unambiguous components of actions, each sub-network can focus exclusively on resolving a single challenge, dense temporal overlaps.

To implement this novel paradigm, we (i) design a network comprising two sub-networks, {\ent} and {\mot}, and (ii) decompose dense action labels into dense action-entity and dense action-motion labels using prompts and a pre-trained large language model (LLM). While both sub-networks receive the same input video, {\ent} focuses solely on detecting dense entity concepts involved in dense actions, whereas {\mot} is dedicated to detecting dense motion concepts involved in dense actions. The dense temporal entity and motion representations learned by the sub-networks are then concatenated for dense action detection. The entire network is optimized using the original dense action labels and the BCE loss, while the {\ent} and {\mot} sub-networks are individually optimized using the dense action-entity and dense action-motion labels with the BCE loss.

In dense action detection, where multiple concepts can occur simultaneously, awareness of class dependencies can significantly enhance performance. For instance, in scenarios like cooking, actions such as \enquote{Pouring} and \enquote{Stirring} often occur together. However, we argue that the current dense action detection networks \cite{tirupattur2021modeling, dai2022ms, sardari2023pat, zhu2024dual} cannot effectively learn the relationships among the co-occurrence classes as they are trained using the BCE loss which treats each action class independently during the optimization process. This limitation motivates us to raise our second novel question: {\bf Can we improve network optimization to fully unlock the potential benefits of co-occurring concepts?} To achieve this, we propose providing explicit supervision on co-occurring concepts in the input video during network optimization. Inspired by contrastive language-image pretraining \cite{radford2021learning}, we introduce Contrastive Co-occurrence Language-Video learning, which aligns the video features in the embedding space with the textual features of all co-occurring classes. Specifically, we assign a textual description to each co-occurring concepts in the input video and use a frozen pre-trained text encoder to extract their features. Then, we adapt the noise contrastive estimation loss to match the video features with the text features of all co-occurring classes. Through this, the network not only receives explicit knowledge of co-occurring concepts during training, but also implicitly benefits from the learned semantics of related concepts within the embedding space of pre-trained language models. 

In Fig. \ref{fig: old vs ours}, we compare current approaches to our proposed method ({\method}) in tackling the dense action detection task.

Our key contributions are summarized as follows: (i) we introduce a paradigm shift in solving dense action detection task—decomposing the problem complexity for the network—an approach that can also benefit solving other dense computer vision problems ({\eg}, dense captioning); (ii) we pioneer the first exploration of explicitly addressing action class ambiguities in dense action detection task; (iii) for the first time, we introduce an optimization process for dense action detection,  which enables the network to leverage explicit supervision on co-occurring concepts during training. This approach can enhance the optimization process of any existing or future dense action detection network; (iv) our comprehensive comparison using multiple metrics and challenging benchmark datasets against state-of-the-art approaches demonstrates the superiority of our method, {\eg}, achieving substantial improvements of {\bf 3.8\%} and {\bf 1.7\%} on average across all metrics on {\char} and {\thum}, respectively, and (v) our extensive ablation studies on these benchmark datasets, evaluated across multiple metrics, highlight the effectiveness of each component in our method’s design.

\section{Related Works}
\label{sec: related works}

\noindent{\bf Dense Action Detection --} Current dense action detection approaches \cite{dai2019tan,tirupattur2021modeling,sardari2023pat,dai2023aan,zhu2024dual} typically follow a common pipeline. First, the video is divided into segments, and a frozen, pre-trained Encoder (e.g., I3D, CLIP) extracts features from each segment. These features are then passed to a Temporal Modeling Block that captures their temporal relationships, followed by a classification layer that maps the learned representations to multi-action probabilities. The network is optimized using BCE loss. Although most of the pipeline is shared across approaches, the primary distinctions lie in the design of the Temporal Modeling Block. Below, we briefly review this block in existing approaches.

Pre-transformer approaches, such as \cite{piergiovanni2018learning, piergiovanni2019temporal, kahatapitiya2021coarse}, rely on Gaussian or convolutional filters to represent a video as a sequence of multi-activity events. While these methods are effective at modeling short, dense actions, the inherent temporal limitations of Gaussian and convolutional kernels restrict their ability to capture longer actions. With the success of transformers in modeling 
 long-term dependencies, several works \cite{tirupattur2021modeling, dai2022ms, sardari2023pat, dai2021pdan, dai2023aan, zhu2024dual} have developed transformer-based networks. Among these, some approaches, such as \cite{dai2022ms, sardari2023pat, tanpointtad, zhu2024dual}, focus on modeling various ranges of temporal relationships using multi-scale transformer networks or DETR-based architectures \cite{carion2020end}. On the other hand, \citet{tirupattur2021modeling} introduce the concept of benefiting from learning co-occurrence class relationships. To learn these relationships, they propose explicitly modeling all action classes within the network architecture. Similarly, \citet{dai2023aan} embed all objects in the dataset into the network’s architecture. However, not only do their designs lack computational efficiency due to their dependence on the maximum number of classes, but they also fail to fully capture co-occurrence relationships despite explicitly modeling the classes, as the networks are still optimized using the BCE loss, which treats each class independently. To the best of our knowledge, for the first time, our proposed contrastive co-occurrence language-video loss, is designed to overcome this limitation in network optimization by providing explicit supervision on co-occurring concepts during training. Furthermore, as it is a general loss function applied in the embedding space, it can benefit the optimization process in any existing or future network.

Although transformer-based approaches show performance improvements over traditional methods, the inherent complexity of handling both temporal and action class overlaps poses a substantial obstacle for networks. We addresses this by eliminating one of the overlaps; we propose to decompose the task of detecting dense ambiguous actions to detecting dense non-ambiguous sub-concepts underlying the action classes, and assign these sub-tasks to distinct sub-networks. By isolating these non-ambiguous components, each sub-network focuses exclusively on resolving a single challenge, dense temporal overlaps.

\noindent{\bf Vision-Language for Action Detection --} Building on CLIP's zero-shot capabilities \cite{radford2021learning}, many works, such as \cite{nag2022zero, li2024detal, fish2024plot, liberatori2024test}, adapt its language-image pre-training paradigm for zero-shot or few-shot action detection. Following this, some works, such as \cite{cao2022locvtp, xu2022contrastive}, explore using language models for network pre-training. In contrast, \cite{ju2023distilling, dai2023aan} integrate language models directly during training. For instance, \citet{dai2023aan} introduce an object-centric graph for indoor activity detection and leverage language supervision to ensure that each graph node corresponds to a distinct object, while \cite{ju2023distilling} use language to obtain pseudo-labels for weakly supervised learning. In a similar spirit, we benefit from language models during training. However, our goal is different from that of prior works; we aim to leverage language to effectively learn the relationships among co-occurring concepts.

\section{Methodology}
\label{sec:method}
In this section, we first define the dense action detection task and briefly review the common pipeline used by current approaches to tackle this task, focusing specifically on the multi-scale transformer-based approach presented in \cite{sardari2023pat}, which serves as the backbone of part of our network. We then elaborate on our proposed approach, {\method}.

\begin{figure*}[t]
  \includegraphics[width=1.0\linewidth]{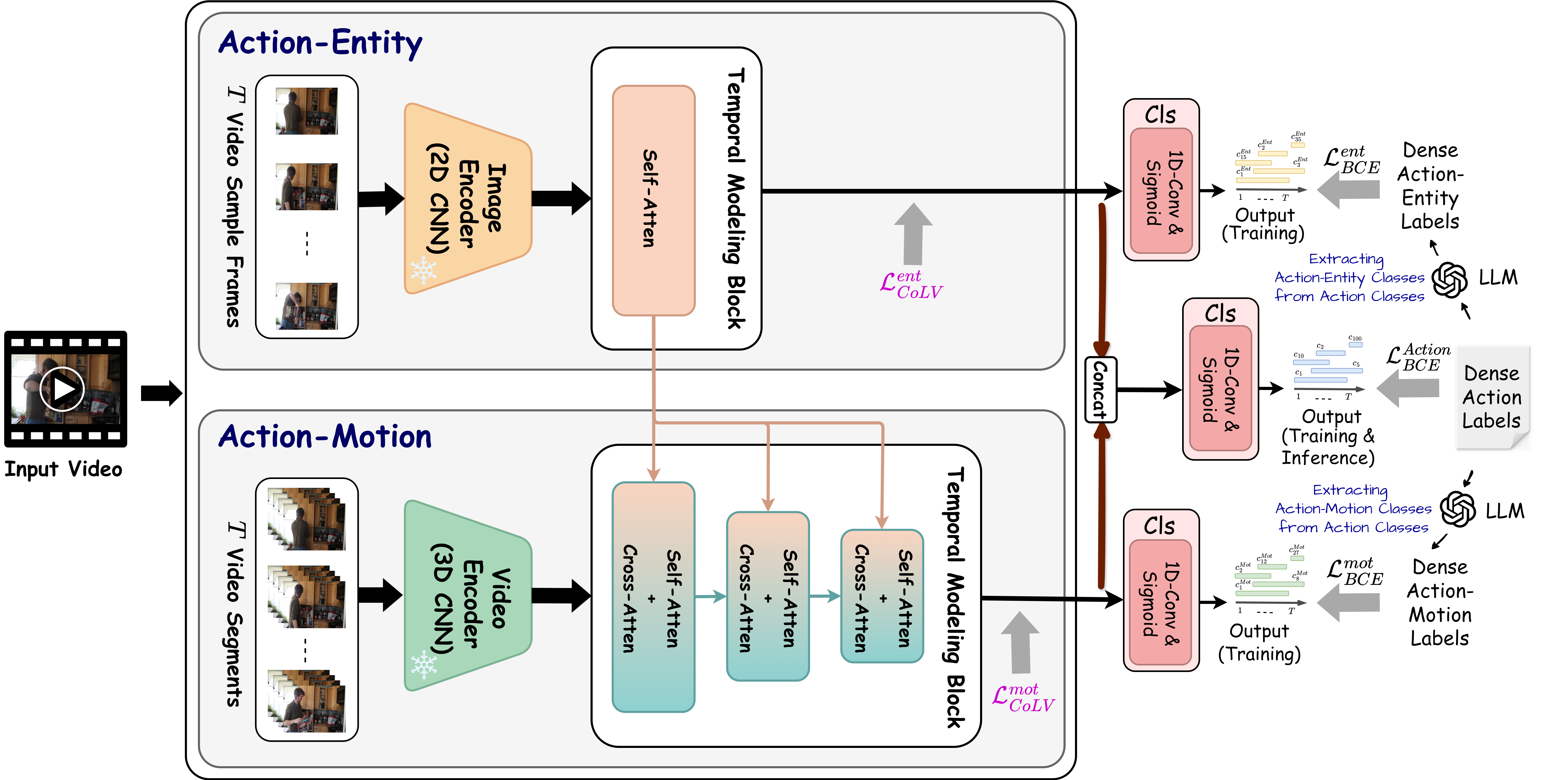}
  \caption{{The overall scheme of {\method}. Our proposed network consists of two sub-networks: {\ent} and {\mot}. {\ent} learns dense entity concepts associated with the action classes, while {\mot} focuses on learning dense motion concepts related to the action classes. The entire network is optimized using the dense action labels and the BCE loss ($\mathcal{L}^{Action}_{BCE}$). Additionally, the sub-networks are optimized using dense action-entity and action-motion labels, which are derived from action labels, along with the BCE loss ($\mathcal{L}^{ent}_{BCE}$, and $\mathcal{L}^{mot}_{BCE}$) and our proposed contrastive co-occurrence language-video loss ($\mathcal{L}^{ent}_{{CoLV}}$ and $\mathcal{L}^{mot}_{{CoLV}}$).} \vspace{-5mm}}
  \label{fig: method}
\end{figure*}

\begin{figure}[t]
  \includegraphics[width=0.8\linewidth]{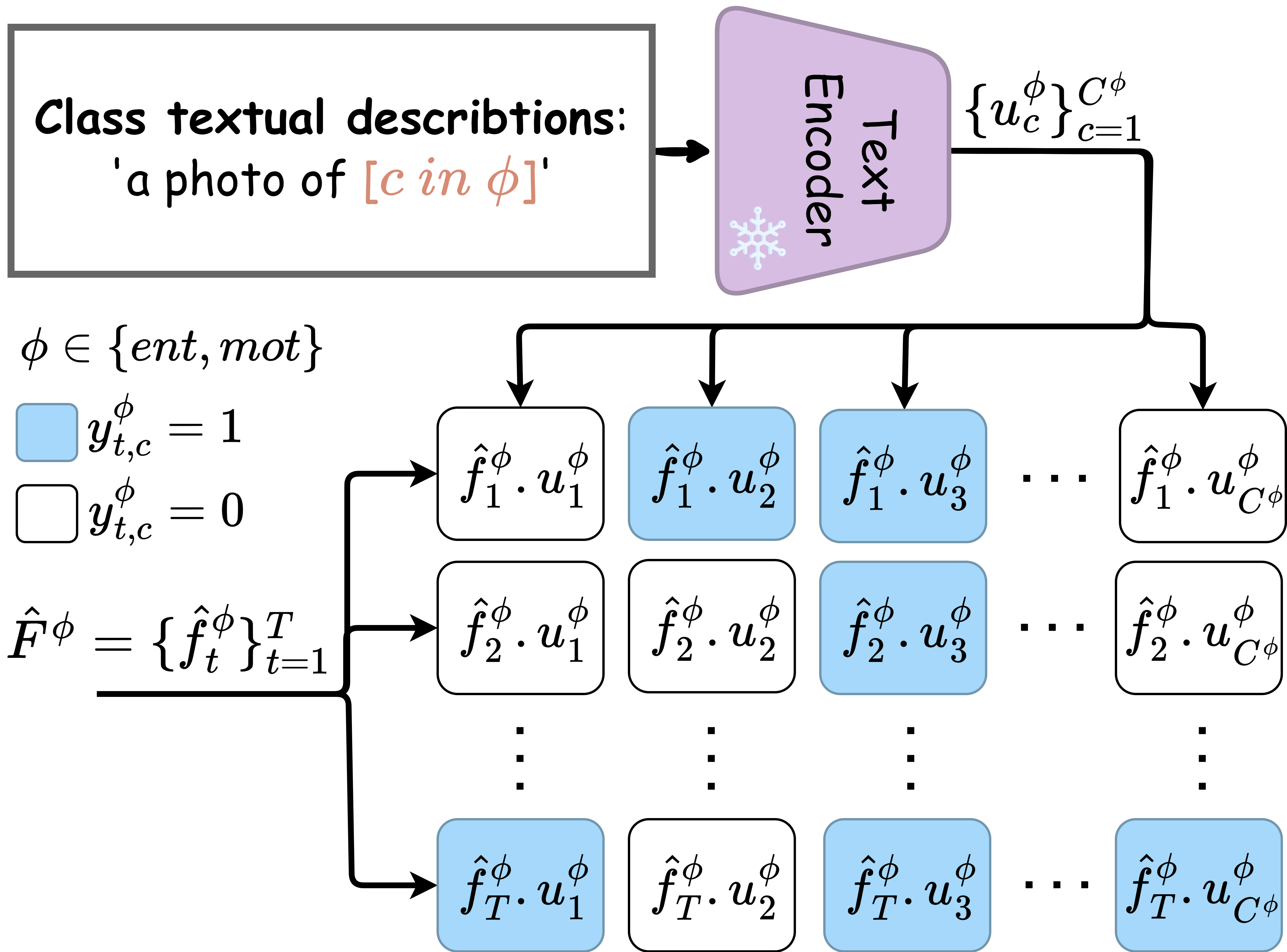}
  \caption{{Alignment of temporal video features with textual features of co-occurring class concepts in our contrastive co-occurrence language-video loss ({\ie}, $\mathcal{L}^{ent}_{{CoLV}}$ and $\mathcal{L}^{mot}_{{CoLV}}$).} }
  \label{fig: loss}
\end{figure}

\subsection{Preliminaries}
\label{sec:preliminaries}
\noindent{\bf Problem Definition --} In the dense action detection task, the goal is to identify all actions occurring at each timestamp of an untrimmed video, as described in \cite{tirupattur2021modeling, dai2022ms, sardari2023pat, zhu2024dual}. Given an untrimmed video sequence $V=\{I_n \in \rm {I\!R}^{W\times H\times 3}\}^{N}_{n=1}$ of length $N$, each timestamp $n$ has a multi-action class label $Y_{n}=\{{{y}_{n,c}}\in\{0, 1\}\}^{C}_{c=1}$, where $C$ represents the total number of action classes in the dataset, and the set of action labels for the entire video is denoted as $Y=\{Y_n\}^{N}_{n=1}$. The network's task is to estimate multi-action class probabilities $P=\{P_n\}^{N}_{n=1}$, where ${P}_n =\{{p}_{n,c}\in[0, 1]\}^{C}_{c=1}$.

\vspace{1.5mm}
\noindent{\bf {Current Pipeline} to Tackle Dense Action Detection --} 
To tackle dense action detection task, the most widely used pipeline consists of three main components: an {Encoder}, a {Temporal Modeling Block}, and a {Classification} layer. The {Encoder}, typically a frozen pre-trained 3D CNN, processes the input video sequence $V$ for the {Temporal Modeling Block}. First, the video is divided into non-overlapping $K$-frame video segments $V=\{S_t\}^{T}_{t=1}$, where $S_t\in \rm {I\!R}^{K\times W\times H\times 3}$ and $T= \frac{N}{K}$. The segments are then fed into the {Encoder} to obtain segment-level input tokens
\begin{equation}
\label{eq:videnc}
    F=\{\text{Encoder}(S_t)\}^T_{t=1},
\end{equation}
\noindent where $F\in \rm {I\!R}^{T\times D}$. The {Temporal Modeling Block} receives the input video tokens to exploit a range of temporal relationships among them. For example, in the recent state-of-the-art method presented in \cite{sardari2023pat}, the {Temporal Modeling Block} is a multi-scale transformer, where a self-attention module is first applied to all input tokens to learn a fine-grained temporal representation of the video
\begin{equation}
\label{eq:fine}
\bar{F}_{fin} =  \text{Self-attn}_{\Theta}({F}), 
\end{equation}
where $\bar{F}_{fin}\in \rm {I\!R}^{T\times D^{\ast}}$. Then, $M$ convolutional layers with different strides are applied to the fine-grained features to down-sample them $\acute{F}_{crs,\theta}=\text{Conv}_{\theta}(\bar{F}_{fin})$, where $\acute{F}_{crs,\theta}\in \rm {I\!R}^{\frac{T}{2^{\theta}}\times D^{\ast}}$ and ${\theta} \in \rm \{1,2,…,M\}$. Subsequently, $M$ self-attention modules are applied to further exploit the temporal dependencies among the coarsely down-sampled features 
\begin{equation}
\label{eq:coarse}
\bar{F}_{crs,\theta} = \text{Self-attn}_{\theta}(\acute{F}_{crs,\theta}),
\end{equation}
where $\bar{F}_{crs,\theta}\in \rm {I\!R}^{\frac{T}{2^{\theta}}\times D^{\ast}}$. The coarse features are then up-sampled through linear interpolation $\bar{\bar{F}}_{crs,\theta}=\text{UpSample}(\bar{F}_{crs,\theta})$, where  $\bar{\bar{F}}_{crs,\theta}\in \rm {I\!R}^{T\times D^{\ast}}$ to match the same temporal length as the original input tokens. The up-sampled features, together with the fine-grained ones, are fused using techniques such as summation or concatenation for action classification $\hat{F} = \text{Fuse}(\bar{F}_{fin},\bar{\bar{F}}_{crs,1}, …, \bar{\bar{F}}_{crs,M})$, where $\hat{F} \in \rm {I\!R}^{T\times D^{\ast}}$. 

\noindent{Finally, the {Classification} layer, typically composed of fully connected or 1D convolutional filters, produces the multi-action class probabilities ${P} = Cls(\hat{F})$, where $P\in \rm {I\!R}^{T\times C}$. The entire network is typically optimized using the ground truth labels $Y$ and the {BCE loss}, $\mathcal{L}_{BCE} = BCE(Y, {P})$ as
\begin{gather}
\footnotesize
    BCE(Y, {P}) = -\frac{1}{T}\sum_{t=1}^{T}\sum_{c=1}^{C}\ell_{\text{bce}}(y_{t,c},p_{t,c}),
\end{gather}
where $\ell_{\text{bce}}(y,p) = ylog(p)+(1-y)log(1-p)$.

\subsection{Reframing Dense Action Detection (RefDense)}
We introduce a paradigm shift in solving the dense action detection task. Instead of tackling the entire complex problem—handling the dual challenge of temporal and action class overlaps (i.e., class ambiguity)—with a single network, we propose decomposing the task into less complex sub-tasks: detecting dense, unambiguous sub-concepts underlying action classes (i.e., entity and motion sub-concepts) and assigning these sub-tasks to distinct sub-networks. By isolating these unambiguous concepts of actions, the sub-networks can focus exclusively on resolving a single challenge—dense temporal overlaps.

\noindent{To implement our proposed paradigm, we (i) design a network comprising two sub-networks: {\ent} and {\mot}, and (ii) decompose dense action labels into dense action-entity and dense action-motion labels using prompts and a pre-trained Large Language Model (LLM), as shown in Fig. \ref{fig: method}. While both sub-networks receive the same input video, {\ent} is tailored to detect dense entity concepts, whereas {\mot} is designed to detect dense motion concepts. The dense temporal entity and motion representations learned by the sub-networks are then concatenated for dense action detection. The entire network is optimized using dense action labels and the BCE loss, while the {\ent} and {\mot} sub-networks are also individually optimized using dense action-entity and dense action-motion labels, respectively, with the BCE loss. Furthermore, to effectively leverage the interrelationships among co-occurring concepts within the video, we optimize the network’s embedding space using our proposed contrastive co-occurrence language-video loss. In the following, we detail our network, label decomposition, and loss functions.}

\vspace{1.5mm}
\noindent{\bf Dense Action-Entity \& Dense Action-Motion Labels --}
These labels are extracted for each input video from its original action labels. First, a set of action-entity and action-motion classes is defined from all action classes using specific prompts and a pre-trained LLM, GPT-4 (see supp. file for details). For example, from the action class \enquote{Weight Lifting Clean}, the action-entity class \enquote{Barbell} and the action-motion class \enquote{Lifting-Clean} are extracted, respectively. Then, for each input video, using its corresponding action ground-truth label $Y$ and the newly defined classes, we generate its dense action-entity and dense motion-entity labels as $Y^{ent}\in \mathbb{R}^{T\times C^{ent}}$ and $Y^{mot}\in \mathbb{R}^{T\times C^{mot}}$, where $ C^{ent}$ and $C^{mot}$ refer to the number of defined action-entity and action-motion classes, and $C^{ent},C^{mot} \leqslant C$. 

\noindent{Note: (i) The decomposed labels preserve the temporal boundaries of the original labels extracted from, and (ii)  not all actions involve both entities and motion components (e.g., \enquote{Walking}). For actions with only one component, the label is generated only for that component.}


\vspace{1.5mm}
\noindent{\bf {\ent} Sub-Network --} This sub-network aims to detect dense entity concepts involved in the action classes. To achieve this, it consists of two components: an {Image Encoder} and a {Temporal Modeling Block}. First, the Image Encoder, a frozen pre-trained 2D CNN, is applied to the sampled frames of the input video $\{I_t\}^{T}_{t=1}$ to extract their spatial features $\mathbb{F}=\{\text{ImageEncoder}(I_t)\}^{T}_{t=1}$, where $I_t \in \rm {I\!R}^{W\times H\times 3}$, and $\mathbb{F}\in \rm {I\!R}^{T\times \mathbb{D}}$. For sampling, the middle frame of each video segment $S_t$ is selected. Then, the {Temporal Modeling Block}, which employs a lightweight transformer, including a few self-attention layers, receives the spatial features to model the dense action-entity concepts.
\begin{equation}
\hat{F}^{ent}= \text{Self-att}_{\Delta}(\mathbb{F}),~\text{where $\hat{F}^{ent} \in \rm {I\!R}^{T\times {D}^{{\ast}}}$.}
\end{equation}

\noindent{\bf {\mot} Sub-Network --} The goal of this sub-network is to detect dense motion concepts involved in the action class. To achieve this, it consists of two components: a {Video Encoder} and a {Temporal Modeling Block}. First, the Video Encoder, a frozen pre-trained 3D CNN, is applied on the video segments to extract their spatio-temporal video features as in Eq. \ref{eq:videnc}, $F=\{\text{VideoEncoder}(S_t)\}^T_{t=1}$. The features are then processed through the {Temporal Modeling Block}. Unlike the dense entity concepts, which can be modeled using only fine-grained temporal video features with a lightweight transformer, learning dense motion is more challenging, as a motion concept can vary in duration across different video samples. For example, the motion concept \enquote{Holding} may last only a few seconds in one video but may take up to a minute in another. To address this, it is necessary to model multiple scales of temporal information. Therefore, to implement the {Temporal Modeling Block} in this sub-network, we use the multi-scale transformer proposed in \cite{sardari2023pat} as backbone. However, in contrast to the original network, which uses only the self-attention mechanism to learn fine and coarse video representations, our {Temporal Modeling Block} benefits additionally from the guidance of the {\ent} sub-network through a cross-attention mechanism. Eq. \ref{eq:fine} and Eq. \ref{eq:coarse} are adapted as
\begin{gather} 
\footnotesize
\bar{F}^{mot}_{fin} = \text{Cross-attn}_{\Theta}(\text{Self-att}_{\Theta}(F), \hat{F}^{ent}),\\ 
\bar{F}^{mot}_{crs,\theta} = \text{Cross-attn}_{\theta}(\text{Self-att}_{\theta}(\acute{F}_{crs,\theta}),\hat{F}^{ent}).
\end{gather}
\noindent This guidance enables the {\mot} sub-network to focus more effectively on regions informed by the learned entity concepts. In term Cross-attn$(a, b)$, the Query is generated from $a$, and the Key and Value are derived from $b$. Finally, similar to the backbone \cite{sardari2023pat}, the dense motion video representations $\hat{F}^{mot}$ are obtained by fusing the fine and coarse motion representations.

\vspace{1.5mm}
\noindent{\bf Sub-Networks Fusion for Dense Action Detection --}
To perform dense action detection, the dense entity and motion video representations learned by the sub-networks are first concatenated to form the full video representation. Then, a 1D convolutional filter is applied to the full features to predict multi-action probabilities for all video segments:
\begin{equation} 
P = \text{Sig}(\text{1D-Conv}_{\vartheta}([\hat{F}^{ent};\hat{F}^{mot}])), 
\end{equation}
where $[;]$ denotes the concatenation operation, $Sig$ refers to the sigmoid activation function, and $P \in \mathbb{R}^{T\times C}$. 

\vspace{1.5mm}
\noindent{\bf Binary Cross-Entropy Optimization --} With the action probabilities ${P}$ and action labels $Y$, the entire network is optimized using $\mathcal{L}^{Action}_{BCE} = BCE(Y, P)$. The {\ent} and {\mot} sub-networks are also individually optimized using BCE and dense action-entity and action-motion labels $Y^{ent}$ and $Y^{mot}$. To perform this, during training, a 1D convolutional layer is added on top of each sub-network to obtain multi-entity and multi-motion probabilities
\begin{gather}
    P^{\phi}=\text{Sig}(\text{1D-Conv}_{\phi}(\hat{F}^{\phi})),
\end{gather}
\noindent where $P^{\phi} \in \mathbb{R}^{N\times C^{\phi}}$ and $\phi \in \{ent, mot\}$. With the probabilities and labels, the network is optimized using $\mathcal{L}^{RD}_{BCE}=\sum_{\phi} \mathcal{L}^{\phi}_{BCE}$, where $\mathcal{L}^{\phi}_{BCE} = BCE(Y^{\phi}, P^{\phi})$.
\vspace{1.5mm}
\noindent{\bf Contrastive Co-Occurrence Language-Video Learning --}
In scenarios where multiple concepts occur simultaneously, awareness of class dependencies can enhance the method’s performance. However, we argue that optimizing with the BCE loss does not allow networks to effectively learn these relationships, as BCE treats each class label independently. To address this limitation, we propose providing explicit supervision on co-occurring concepts in the video during training. To achieve this, inspired by contrastive language-image pre-training \cite{radford2021learning}, we align the learned video representations in the embedding space $\hat{F}^{\phi}=\{\hat{f}^{\phi}_{t}\}^T_{t=1}$ with the extracted text features of all co-occurring classes in the input video (see Fig. \ref{fig: loss}). {Specifically, a textual sentence is assigned to each class occurring in the video as $txt^{\phi}_{c}=\text{`a photo of }[c\text{ in }{\phi}]\text{'}$, where $[c\text{ in }{\phi}]$ represents the text description for class $c$ within the class set $\phi$, and $\phi \in \{ent, mot\}$.} Then, a frozen pre-trained {Text Encoder} ({\eg}, CLIP’s text encoder) is then used to extract their features $u^{\phi}_{c}=\text{TextEncoder}(txt^{\phi}_{c})$. Finally, the noise contrastive estimation is adapted to match the visual representations of each video segment with the text features of all the co-occurring concepts in that segment as:

\begin{equation}
    \mathcal{L}^{RD}_{{CoLV}} = \sum_{\phi} \mathcal{L}^{\phi}_{{CoLV}},
\end{equation}
\begin{equation}
\small
    \mathcal{L}^{\phi}_{CoLV} = - \frac{1}{T} * \sum_{t=1}^{T}\frac{1}{|\beta(t)^{\phi}|} \hspace{-0.5em}\sum_{e \in \beta(t)^{\phi}} \hspace{-0.5em}\log{\frac{\exp{({{\hat{f}^{\phi}}_{t}}{}^{\intercal}.{u}^{\phi}_{e}/\tau)}}{\sum_{\substack{c=1,\\c \notin \beta(t)^{\phi}}}^{C^{\phi}}\exp{({{\hat{f}^{\phi}}_{t}}{}^{\intercal}.{u}^{\phi}_{c}/\tau)}}},
\end{equation}
\begin{equation}
\small
\beta(t)^{\phi} = \{ e \mid e \in \{1,2,..., C^{\phi}\}, y^{\phi}_{t,e}=1 \}.
\end{equation}

\noindent Through this, the network not only receives explicit knowledge of co-occurring concepts, but also implicitly benefits from the learned semantic among related concepts within the embedding space of pre-trained language models.

\section{Experimental Results}
\label{sec:experiments}
\noindent{\bf Datasets --} 
We evaluate our proposed approach on the primary benchmark datasets for this task, {\char} \cite{charades} and {\thum} \cite{multithomus}, as well as on the recent TSU dataset \cite{dai2022toyota}, with results and details included in the supp. file. {\char} is a large-scale dataset containing 9,848 videos of daily activities across 157 action classes, with a high degree of temporal overlap among action instances. {\thum}, the dense multi-label version of the single-label action detection dataset THUMOS'14 \cite{jiang2014thumos}, includes 413 long sports activity videos across 65 action classes. {\char} and {\thum} considered very challenging datasets as with the current state-of-the-art mean Average Precision (mAP) reaching only 32.0\% and 45.5\%.



\noindent{\bf Implementation Details --}
For the {Video Encoder}, following previous works \cite{tirupattur2021modeling, kahatapitiya2021coarse, sardari2023pat}, we use the pre-trained I3D network \cite{carreira2017quo}. For the {Image Encoder} and {Text Encoder}, we employ CLIP’s pre-trained ResNet-50 image encoder and CLIP’s text encoder \cite{radford2021learning}. In Action-Entity, we implement the {Temporal Modeling Block} using one position-aware self-attention block as designed in \cite{sardari2023pat}. In Action-Motion, we utilize the multi-scale transformer PAT \cite{sardari2023pat} as the backbone for the {Temporal Modeling Block}. For more details on the network architecture, see the Supp. file. The length of each video segment is set to $K=8$ frames. During training, $T=256$ consecutive video segments are randomly sampled from an untrimmed video sequence to serve as network input. At inference, we follow previous work \cite{tirupattur2021modeling, kahatapitiya2021coarse, sardari2023pat} and make predictions on the full video sequence. For {\char}, 38 action-entity and 38 action-motion classes are defined, while for {\thum}, 28 action-entity and 50 action-motion classes are defined.


We conducted our experiments using PyTorch on an NVIDIA GeForce RTX 3090 GPU. Our model was trained with the Adam optimizer, starting with an initial learning rate of 0.0001. We used a batch size of 5 for 25 epochs and a batch size of 3 for 300 epochs and for {\char} and {\thum}, respectively. The learning rate was reduced by a factor of 10 after every 7 epochs for {\char} and after every 130 epochs for {\thum}. Note that the different training settings for {\char} and {\thum} are due to their varying sizes. 

\subsection{State-of-the-Art Comparison}
In this section, we compare the performance of our approach with current state-of-the-art methods using different metrics. {Note: Here, our results and comparisons are based on RGB input features. However, results and comparisons incorporating RGB and optical flows can be found in the supp. file.} 
\begin{table}[t]
\scalebox{0.95}
{
  \centering
  \setlength{\tabcolsep}{2pt} 
    \renewcommand{\arraystretch}{0.7} 
  \begin{tabular}{@{}l l c c c@{}}\specialrule{.2em}{.1em}{.1em}
    \multicolumn{2}{l}{\multirow{2}{*}{Method}}& \multirow{2}{*}{\footnotesize{GFLOPs}}&\multicolumn{2}{c}{mAP(\%)}\\ \cmidrule{4-5}
          &  & &\multicolumn{1}{c}{\footnotesize{\char}} & \multicolumn{1}{c}{\footnotesize{\thum}}\\ \specialrule{.2em}{.1em}{.1em}
          R-C3D \cite{xu2017r}&\footnotesize{ICCV 2017}& -&12.7 & - \\ 
          SuperEvent \cite{piergiovanni2018learning}&\footnotesize{CVPR 2018}& 0.8&18.6&36.4\\ 
          TGM \cite{piergiovanni2019temporal}&\footnotesize{ICML 2019}&1.2&20.6&37.2\\ 
          PDAN \cite{dai2021pdan}&\footnotesize{WACV 2021}&3.2&23.7&40.2\\ 
          CoarseFine \cite{kahatapitiya2021coarse}&\footnotesize{CVPR 2021}&- & 25.1&-\\ 
          MLAD \cite{tirupattur2021modeling}&\footnotesize{CVPR 2021}& 44.8&18.4&42.2\\ 
          CTRN \cite{dai2021ctrn}&\footnotesize{BMVC 2021} & -&25.3& {{44.0}}\\
          PointTAD \cite{tanpointtad}&\footnotesize{NeurIPS 2022} &-&21.0 & 39.8 \\
          MS-TCT \cite{dai2022ms}&\footnotesize{CVPR 2022}& 6.6&{{25.4}}&{43.1}\\ 
          {PAT \cite{sardari2023pat}}& \footnotesize{ICCVW 2023}&8.5&{26.5}&{44.6} \\ 
          TTM \cite{ryoo2023token} & \footnotesize{CVPR 2023} &-& 28.8 & - \\
          ANN \cite{dai2023aan} & \footnotesize{BMVC 2023} &- & \underline{32.0} & - \\
          DualDET \cite{zhu2024dual} &\footnotesize{CVPR 2024} & 5.5 &23.2& \underline{45.5}\\ \midrule
          \multicolumn{2}{l}{\bf{\method}}&11.5&{\bf 33.4}&{\bf 46.6} \\ 
           & & &\blue{(+1.4)} & \blue{(+1.1)}\\ \specialrule{.2em}{.1em}{.1em}
  \end{tabular}}
  \caption{Dense action detection results on the {\char} and {\thum} datasets using RGB inputs, in terms of per-frame mAP. The best and the second best results are in {\bf Bold} and \underline{underlined}.} 
  \label{tab:sota general}
\end{table}

\begin{table*}[t]
\scalebox{0.90}
{
  \centering
  \setlength{\tabcolsep}{1.0pt} 
    \renewcommand{\arraystretch}{0.7} 
      \begin{tabular}{@{}l c c c c c c c c c c c c c c c c c c c@{}} \specialrule{.2em}{.1em}{.1em}
          \multicolumn{1}{l}{\multirow{3}{*}{Method}} & \multicolumn{9}{c}{\char} &~~~~& \multicolumn{9}{c}{\thum}\\ \cmidrule{2-10} \cmidrule{12-20}
          & \multicolumn{4}{c}{\bf {$\tau = 0$}}&&\multicolumn{4}{c}{\bf {$\tau = 20$}}&& \multicolumn{4}{c}{\bf {$\tau = 0$}}& &\multicolumn{4}{c}{\bf {$\tau = 20$}}\\ \cmidrule{2-5} \cmidrule{7-10} \cmidrule{12-15} \cmidrule{17-20}
          &{{mAP}$_{ac}$} & {{F1}$_{ac}$} &{P$_{ac}$}&{R$_{ac}$}& &{{mAP}$_{ac}$} & {{F1}$_{ac}$} &{P$_{ac}$}&{R$_{ac}$}& &{{mAP}$_{ac}$} & {{F1}$_{ac}$} &{P$_{ac}$}&{R$_{ac}$}& &{{mAP}$_{ac}$} & {{F1}$_{ac}$} &{P$_{ac}$}&{R$_{ac}$}\\ \specialrule{.2em}{.1em}{.1em}
          MLAD \cite{tirupattur2021modeling}&28.4 &12.5 &21.7 &8.6& & 34.7& 13.6& 21.0& 10.1 && 18.0& 29.4& 28.8& 13.0& & 19.6& 30.5& 31.4&14.2 \\
          CTRN \cite{dai2021ctrn} &29.7 &11.9 &23.9 &8.1& & 36.8& 12.9& 27.1& 9.1 && -& -& -& -& & -& -& -&- \\
          MS-TCT \cite{dai2022ms}&29.4 &14 &24.8 &9.7& & 35.1& 15.4& 24.3& 11.1 &&  26.3&33.5 &\underline{33.8} &21.5 & &28.8 & 35.4& \underline{37.8}& 22.8\\ 
          {PAT \cite{sardari2023pat}} &{30.0} &\underline{27.1} &{25.9} &\underline{28.4}& & {36.3}& \underline{30.2}& {28.9}& \underline{31.7} &&\underline{29.1}  & \underline{35.0}& {33.5}& \underline{25.7}& & \underline{31.4}& \underline{37.5}& 37.3&\underline{27.1} \\
          {ANN \cite{dai2023aan}} &\underline{35.4} &{20.4} &\underline{31.4} &-& & \underline{41.8}& {22.3}& \underline{30.4}& {-} &&-&-& -& -& &-& -& -&- \\ \midrule
          {\bf {\method}} &\bf 37.5 &\bf 33.0 &\bf 32.0 &\bf 33.9& & \bf 43.7& \bf 36.4&\bf 35.4&\bf 37.4 &&\bf 31.1& \bf 37.1& \bf 35.5& \bf 27.7& & \bf 33.1& \bf 39.2& \bf 38.8&\bf 28.8 \\ 
          & \blue{(+2.0)}& \blue{(+5.9)}& \blue{(+0.6)}&\blue{(+5.5)}& & \blue{(+1.9)}& \blue{(+6.2)}& \blue{(+5.0)}&\blue{(+5.7)}&&\blue{(+2.0)}&\blue{(+2.1)}&\blue{(+1.7)}&\blue{(+2.0)}& &\blue{(+1.7)}& \blue{(+1.7)}& \blue{(+1.0)}& \blue{(+1.7)}\\ \specialrule{.2em}{.1em}{.1em}
    \end{tabular}
  }
  \caption{Dense action detection results on {\char} and {\thum} using RGB inputs, evaluated based on the action-conditional metrics with cross-action dependencies over a temporal window of size $\tau$. The best and the second best results are in {\bf Bold} and \underline{underlined}.} 
  \label{tab:sota conditional}
\end{table*}

The primary metric for dense action detection task is the standard per-frame mAP. Table \ref{tab:sota general} presents comparative results on {\char} and {\thum} using this metric. The results demonstrate the superiority of our approach over state-of-the-art methods, achieving significant improvements of {\bf 1.4\%} and {\bf 1.1\%} mAP on {\char} and {\thum}, respectively, which corresponds to a substantial relative improvement of {\bf 4.4\%} and {\bf 2.5\%} over the current best-performing methods. Furthermore, the results reveal that our approach exhibits better generalization across different datasets, with the smallest performance difference between the two datasets (13.2\%). In contrast, other methods are less consistent; for example, DualDET \cite{zhu2024dual} shows a gap of 22.3\%, and ANN \cite{dai2023aan} is limited to indoor activity datasets due to its object-centric architecture.


The standard mAP assesses the performance by evaluating each class independently. However, it does not explicitly measure whether models learn the relationships amongst the classes. To overcome this, \cite{tirupattur2021modeling} introduce a set of action-conditional metrics, including action-conditional mean Average Precision {{mAP}$_{ac}$}, action-conditional F1-Score {{F1}$_{ac}$}, action-conditional Precision {P$_{ac}$}, and action-conditional Recall {R$_{ac}$}. These metrics aim to explicitly assess how well pairwise class/action dependencies are modeled, both within a single frame and across different frames. Table \ref{tab:sota conditional} presents the comparative results on {\char} and {\thum} using action-conditional metrics. While these metrics evaluate a method’s performance more effectively than standard mAP, only a few methods report results using them, primarily on {\char}. Therefore, for a comprehensive comparison, we produced the results of previous methods under these metrics, using RGB inputs, with their publicly available code whenever accessible. We hope this comprehensive comparison on two datasets benefits the community in future works. Table \ref{tab:sota conditional} demonstrates the superiority of our method over current state-of-the-art approaches in detecting dense actions. Specifically, it achieves an average improvement of {\bf 4.1\%} on {\char} and {\bf 1.8\%} on {\thum} across all conditional metrics.

\noindent{\bf {Qualitative Comparison --}}
In Fig. \ref{fig: qualitative}, we qualitatively compare our approach with the state-of-the-art methods PAT \cite{sardari2023pat} and MS-TCT \cite{dai2022ms} on a test video sample of {\char}. The results show that not only do the action predictions of our method have better overlap with the ground truth labels than the other methods, but it also detects more action classes, {\ie}, Our method detects 4 action types, while PAT and MS-TCT detect 3 and 2 action types, respectively.

\begin{figure}[t]
  \includegraphics[width=1.0\linewidth]{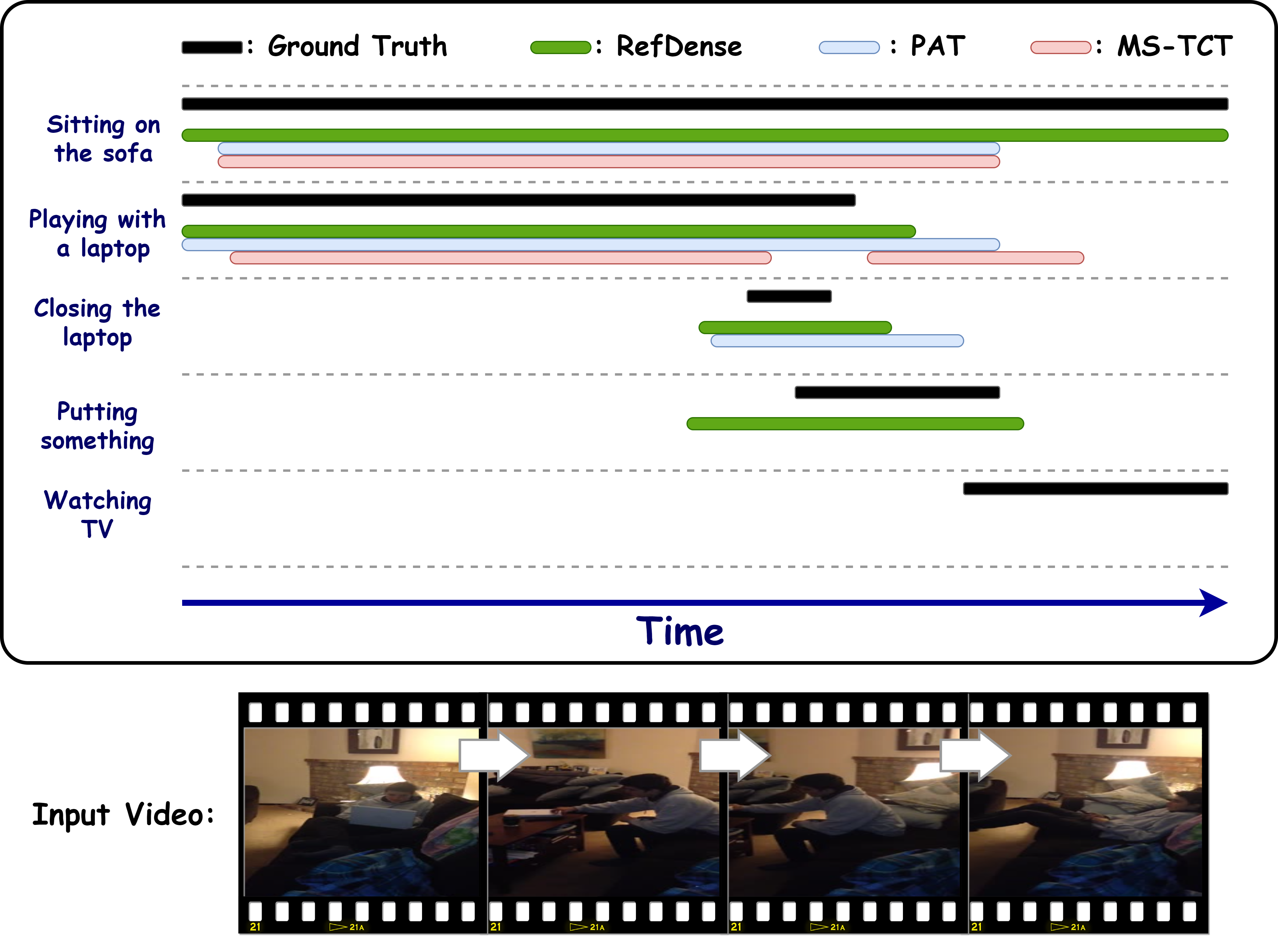}
  \caption{Qualitative comparison with previous approaches (PAT \cite{sardari2023pat} and MS-TCT \cite{dai2022ms}) on a test video sample of {\char}.}
  \label{fig: qualitative}
\end{figure}
\subsection{Ablation Studies}
In this section, we extensively evaluate the impact of key components of our proposed approach using both type of metrics and on two datasets. Note, to perform these experiments all action conditional metrics are measured over a temporal window of size $\tau=0$.


\noindent{\bf {How Sub-Networks Tackle the Task Independently -- }} Table \ref{tab: decomposing} compares the performance of our approach with its individual sub-networks, each challenged to independently solve the entire complex problem (\ie, handing both temporal and class overlaps). To ensure a fair comparison, we adapted these baselines to utilize I3D and CLIP image features through feature concatenation. The findings in Table \ref{tab: decomposing} demonstrates that our approach significantly surpasses its sub-networks across both datasets and all metrics. Notably, it surpasses Action-Entity by an average of {\bf 5.6\%} on {\char} and {\bf 2.9\%} on {\thum}, and Action-Motion by {\bf 4.7\%} on {\char} and {\bf 1.3\%} on {\thum}. 

\begin{table}[t]
\scalebox{0.95}{
  \centering
  \setlength{\tabcolsep}{1.3pt} 
    \renewcommand{\arraystretch}{0.8} 
  \begin{tabular}{@{}l c c c c c c c @{}}\specialrule{.2em}{.1em}{.1em}
    \multirow{2}{*}{Network}& \multicolumn{3}{c}{{\char}} &~~~& \multicolumn{3}{c}{{\thum}} \\ \cmidrule{2-4} \cmidrule{6-8}
    & mAP & {{mAP}$_{ac}$} & {{F1}$_{ac}$}  &&mAP & {{mAP}$_{ac}$} & {{F1}$_{ac}$}\\\specialrule{.2em}{.1em}{.1em}
    {Action-Entity} & 27.7& 31.2& 28.4& & 40.0 & 28.5& 32.3 \\ 
    & \blue{(-5.7)} & \blue{(-6.3)} & \blue{(-4.6)} & & \blue{(-6.6)} & \blue{(-2.6)} & \blue{(-4.8)} \\ \cmidrule{2-8}
    {Action-Motion} & 30.4& 35.0& 30.9& & 44.4& 30.6& 36.0 \\ 
    & \blue{(-3.0)} & \blue{(-2.5)} & \blue{(-3.1)} & & \blue{(-2.2)} & \blue{(-0.5)} & \blue{(-1.1)}  \\ \cmidrule{1-8}
    {\bf {\method}} & \bf 33.4 & \bf 37.5 & \bf 33.0& & \bf 46.6 & \bf 31.1 & \bf 37.1 \\
   \specialrule{.2em}{.1em}{.1em} 
  \end{tabular}
  \caption{Ablation studies on network design.} 
  \label{tab: decomposing}
  }
\end{table}

\vspace{1.0mm}
\noindent{\bf Impact of {Sub-Labels} -- }
The goal of decomposed labels, dense action-entity and action-motion labels, is to ensure that each sub-network focuses solely on its own sub-task. The results in Table \ref{tab: labels} confirm their essential role in addressing the problem. Removing these labels from training leads to a significant performance drop, with an average decrease of {\bf 2.6\%} on {\char} and {\bf 1.6\%} on {\thum}.


\begin{table}[t]
\scalebox{0.92}{
  \centering
  \setlength{\tabcolsep}{1.5pt} 
    \renewcommand{\arraystretch}{0.8} 
  \begin{tabular}{@{}c c c c c c c c c c c @{}}\specialrule{.2em}{.1em}{.1em}
    \multicolumn{2}{c}{Labels}& & \multicolumn{3}{c}{{\char}} & & \multicolumn{3}{c}{{\thum}}\\ \cmidrule{1-2} \cmidrule{4-6} \cmidrule{8-10}
    {Entity} & {Motion}& & {mAP} & {{mAP}$_{ac}$}&{{F1}$_{ac}$}& &{mAP}&{{mAP}$_{ac}$} & {{F1}$_{ac}$} \\\specialrule{.2em}{.1em}{.1em}
    \xmark& \cmark & & 31.1& 34.4& 30.8&& 44.8& 30.8& 36.4\\ 
    \cmark& \xmark & & 31.5& 35.4& 31.3& & 44.6& 29.8& 36.2 \\ \cmidrule{4-10} \cmidrule{4-10} \cmidrule{4-10} \cmidrule{4-10} \cmidrule{4-10} \cmidrule{4-10} \cmidrule{1-9} \cmidrule{1-9}
    \xmark& \xmark && 30.9 & 34.5& 30.9& & 44.5 &29.8 & 36.0 \\ \cmidrule{4-10}
    \cmark& \cmark && \bf 33.4 & \bf 37.5 & \bf 33.0& & \bf 46.6 & \bf 31.1 & \bf 37.1 \\ 
    & & &\blue{(+2.5)} & \blue{(+3.0)} & \blue{(+2.1)} & & \blue{(+2.1)} & \blue{(+1.3)} & \blue{(+1.1)} \\ \specialrule{.2em}{.1em}{.1em}
  \end{tabular}
  \caption{Ablation studies on employing sub-labels for training.} 
  \label{tab: labels}
  }
\end{table}

\vspace{1.0mm}
\noindent{\bf Impact of \texorpdfstring{$\mathcal{L}^{RD}_{CoLV}$}{L^RD_CoLV} --} We ablate our proposed contrastive co-occurrence language-video loss $\mathcal{L}^{RD}_{CoLV}$ in Table \ref{tab: loss}. The results indicate that providing explicit supervision on co-occurring concepts through our loss significantly enhances the method’s performance, over {\bf 1.0\%} improvement, across all metrics. Notably, this improvement is achieved purely through optimization, without modifying the network. 


\noindent{\bf Generalization of \texorpdfstring{$\mathcal{L}^{RD}_{CoLV}$}{L^RD_CoLV} --}
Our proposed loss, $\mathcal{L}^{RD}_{CoLV}$, is a general loss function that can be applied to the embedding space of any existing or future network to improve their optimization. For, example, in Table \ref{tab: gen-loss}, we present its impact when applied to the existing PAT network \cite{sardari2023pat}. The results show that it enhances PAT’s performance across all metrics. Specifically, the standard mAP and {mAP}$_{ac}$ increase significantly, by {\bf 1.1\%} on {\char} and {\thum}, respectively. This improvement is achieved in an end-to-end manner without altering the network architecture.

\vspace{1.0mm}
\noindent{\bf Impact of Cross-Attention Mechanism --}
To enhance the learning of motion concepts, we design the Action-Motion sub-network to receive guidance from the Action-Entity sub-network through the cross-attention mechanism, allowing it to focus more on regions highlighted by the learned action-entity concepts. The results in Table \ref{tab: cross-attention} verify the effectiveness of this design, showing that by adding the cross-attention mechanism, we achieve over {\bf 1.0\%} improvement across most metrics on both datasets.

\begin{table}[t]
\scalebox{0.95}{
  \centering
  \setlength{\tabcolsep}{3pt} 
    \renewcommand{\arraystretch}{0.8} 
  \begin{tabular}{@{}c c c c c c c c c @{}}\specialrule{.2em}{.1em}{.1em}
    \multirow{2}{*}{$\mathcal{L}^{RD}_{CoLV}$} & \multicolumn{3}{c}{{\char}} & & \multicolumn{3}{c}{{\thum}} \\ \cmidrule{2-4} \cmidrule{6-8}
    & {mAP}& {{mAP}$_{ac}$} & {{F1}$_{ac}$} && {mAP}& {{mAP}$_{ac}$} & {{F1}$_{ac}$}\\\specialrule{.2em}{.1em}{.1em}
    \xmark & 32.2& 36.2& 32.0& & 44.9 & 30.1& 35.9 \\ \cmidrule{2-8} 
    \cmark& \bf 33.4 & \bf 37.5 & \bf 33.0& & \bf 46.6 & \bf 31.1 & \bf 37.1\\ 
    & \blue{(+1.2)} & \blue{(+1.3)} & \blue{(+1.0)} & & \blue{(+1.7)} & \blue{(+1.0)} & \blue{(+1.2)}\\
   \specialrule{.2em}{.1em}{.1em}
  \end{tabular}
  \caption{Ablation studies on $\mathcal{L}^{RD}_{CoLV}$.} 
  \label{tab: loss}
  }
\end{table}

\begin{table}[t]
\scalebox{0.95}{
  \centering
  \setlength{\tabcolsep}{1.pt} 
    \renewcommand{\arraystretch}{0.9} 
  \begin{tabular}{@{}l c c c c c c c @{}}\specialrule{.2em}{.1em}{.1em}
    & \multicolumn{3}{c}{{\char}} & & \multicolumn{3}{c}{\thum} \\ \cmidrule{2-4} \cmidrule{6-8}
    {} & {mAP}& {{mAP}$_{ac}$} & {{F1}$_{ac}$} && {mAP}& {{mAP}$_{ac}$} & {{F1}$_{ac}$}\\\specialrule{.2em}{.1em}{.1em}
    \footnotesize{PAT \cite{sardari2023pat}} & 25.6& 30.0& 27.1& & 44.6 & 29.1& 35.0\\ \cmidrule{2-8} 
    \footnotesize{PAT \cite{sardari2023pat} + $\mathcal{L}^{RD}_{CoLV}$}& \bf 26.7 & \bf 30.7 & \bf 27.7& & \bf 45.2 & \bf 30.2 & \bf 35.4\\ 
    & \blue{(+1.1)} & \blue{(+0.7)} & \blue{(+0.6)} & & \blue{(+0.6)} & \blue{(+1.1)} & \blue{(+0.4)}  \\
   \specialrule{.2em}{.1em}{.1em} 
  \end{tabular}
  \caption{Impact of $\mathcal{L}^{RD}_{CoLV}$ on PAT \cite{sardari2023pat}.} 
  \label{tab: gen-loss}
  }
\end{table}

\begin{table}[t]
\scalebox{0.95}{
  \centering
  \setlength{\tabcolsep}{1.6pt} 
    \renewcommand{\arraystretch}{0.8} 
  \begin{tabular}{@{}c  c c c c c c c c @{}}\specialrule{.2em}{.1em}{.1em}
   \multirow{2}{*}{Cross-Attn}&\multicolumn{3}{c}{{\char}} & & \multicolumn{3}{c}{{\thum}} \\ \cmidrule{2-4} \cmidrule{6-8}
    & mAP & {{mAP}$_{ac}$} & {{F1}$_{ac}$} && mAP & {{mAP}$_{ac}$} & {{F1}$_{ac}$}\\\specialrule{.2em}{.1em}{.1em}
    \xmark& 31.8& 35.8& 31.4& &44.7& 29.8& 36.4 \\ \cmidrule{2-8}
    \cmark& \bf 33.4 & \bf 37.5 & \bf 33.0& &\bf 46.6 & \bf 31.1 & \bf 37.1\\ 
    &\blue{(+1.6)}& \blue{(+1.7)}&\blue{(+1.6)}&&\blue{(+1.9)}& \blue{(+1.3)}&\blue{(+0.7)}\\ 
    \specialrule{.2em}{.1em}{.1em}
  \end{tabular}
  \caption{Ablation studies on using the cross-attention mechanism. } 
  \label{tab: cross-attention}
  }
\end{table}


\section{Conclusion}
In this paper, we introduce a paradigm shift in solving the dense action detection task. Instead of tackling the entire complex problem—handling the dual challenge of temporal and action class overlaps ({\ie}, class ambiguity)—using a single network, we propose decomposing the task of detecting dense, ambiguous actions into detecting dense, unambiguous sub-concepts that define the action classes, and assigning these sub-tasks to distinct sub-networks. By isolating these unambiguous concepts, each sub-network can focus exclusively on resolving a single challenge—dense temporal overlaps. Furthermore, to effectively learn the relationships among co-occurring concepts in a video, we propose a novel contrastive language-guided loss that provides explicit supervision on co-occurring concepts during training. Our extensive experiments, conducted on the challenging benchmark datasets {\char} and {\thum} using multiple metrics, demonstrate that our method significantly outperforms state-of-the-art approaches across all metrics. Additionally, ablation studies highlight the effectiveness of the key components of our method. Future work will extend our approach to dense multi-modal ({\eg}, audio-visual) dense action detection.
{
    \small
    \bibliographystyle{ieeenat_fullname}
    \bibliography{main}
}


\end{document}